\documentclass[sn-mathphys-num]{sn-jnl}% Math and Physical Sciences Numbered 
\usepackage{graphicx}%
\usepackage{multirow}%
\usepackage{amsmath,amssymb,amsfonts}%
\usepackage{amsthm}%
\usepackage{mathrsfs}%
\usepackage[title]{appendix}%
\usepackage{xcolor}%
\usepackage{textcomp}%
\usepackage{manyfoot}%
\usepackage{booktabs}%
\usepackage{algorithm}%
\usepackage{algorithmicx}%
\usepackage{algpseudocode}%
\usepackage{listings}%
\usepackage[nolist]{acronym}
\usepackage{comment}
\usepackage{makecell}
\usepackage{hyperref}
\usepackage{xr}
\makeatletter
\newcommand*{\addFileDependency}[1]{% argument=file name and extension
  \typeout{(#1)}% latexmk will find this if $recorder=0 (however, in that case, it will ignore #1 if it is a .aux or .pdf file etc and it exists! if it doesn't exist, it will appear in the list of dependents regardless)
  \@addtofilelist{#1}% if you want it to appear in \listfiles, not really necessary and latexmk doesn't use this
  \IfFileExists{#1}{}{\typeout{No file #1.}}% latexmk will find this message if #1 doesn't exist (yet)
}
\makeatother

\newcommand*{\myexternaldocument}[1]{%
    \externaldocument{#1}%
    \addFileDependency{#1.tex}%
    \addFileDependency{#1.aux}%
}
\myexternaldocument{supp}

\newcommand{\valunit}[2]{\ensuremath{{#1\,\mathrm{#2}}}}

\theoremstyle{thmstyleone}%
\theoremstyle{thmstyletwo}%

\theoremstyle{thmstylethree}%

\begin{document}
%TC:ignore
\begin{acronym}
    \acro{MAC}{Multipy-Accumulate}
    \acro{IMC}{In-Memory-Computing}
    \acro{LLM}{Large Language Model}
    \acro{DRAM}{dynamic Random Access Memory}
    \acro{ASIC}{Application-Specific Integrated Circuit}
    \acro{FPGA}{Field-Programmable Gate Array}
    \acro{PWM}{Pulse-Width Modulation}
    \acro{DAC}{Digital to Analog Converter}
    \acrodefplural{DAC}[DACs]{Digital to Analog Converters}
    \acro{ADC}{Analog to Digital Converter}
    \acrodefplural{ADC}[ADCs]{Analog to Digital Converters}
    \acro{IGZO}{Indium Gallium Zinc Oxide}
    \acro{NLP}{Natural Language Processing}
    \acro{HBM}{High Bandwidth Memory}
    \acro{GPU}{Graphical Processing Units}
    \acro{OSFET}{Oxide Semiconductor Field Effect Transistor}
    \acro{ITO}{Indium Tin Oxide }
\end{acronym}

\title[Article Title]{Analog In-Memory Computing Attention Mechanism for Fast and Energy-Efficient Large Language Models}

\author*[1]{\fnm{Nathan} \sur{Leroux}}\email{n.leroux@fz-juelich.de}
\equalcont{These authors contributed equally to this work.}

\author*[2,3]{\fnm{Paul-Philipp} \sur{Manea}}\email{p.manea@fz-juelich.de}
\equalcont{These authors contributed equally to this work.}

\author[2]{\fnm{Chirag} \sur{Sudarshan}}
\author[1,3]{\fnm{Jan} \sur{Finkbeiner}}
\author[2]{\fnm{Sebastian} \sur{Siegel}}
\author[2,3]{\fnm{John Paul} \sur{Strachan}}
\author[1,3]{\fnm{Emre} \sur{Neftci}}

\affil[1]{\orgdiv{PGI-15}, \orgname{Forschungszentrum J\"ulich}, \orgaddress{\city{J\"ulich}, \country{Germany}}}

\affil[2]{\orgdiv{PGI-14}, \orgname{Forschungszentrum J\"ulich}, \orgaddress{\city{J\"ulich}, \country{Germany}}}

\affil[3]{\orgdiv{Faculty of Electrical Engineering} \orgname{RWTH Aachen}, \orgaddress{\city{Aachen}, \country{Germany}}}

\abstract{Transformer networks, driven by self-attention, are central to Large Language Models. In generative Transformers, self-attention uses cache memory to store token projections, avoiding recomputation at each time step. However, GPU-stored projections must be loaded into SRAM for each new generation step, causing latency and energy bottlenecks.

We present a custom self-attention in-memory computing architecture based on emerging charge-based memories called gain cells, which can be efficiently written to store new tokens during sequence generation and enable parallel analog dot-product computation required for self-attention. However, the analog gain cell circuits introduce non-idealities and constraints preventing the direct mapping of pre-trained models.
To circumvent this problem, we design an initialization algorithm achieving text processing performance comparable to GPT-2 without training from scratch. Our architecture respectively reduces attention latency and energy consumption by up to two and five orders of magnitude compared to GPUs, marking a significant step toward ultra-fast, low-power generative Transformers.
}


\maketitle
%TC:endignore
\section{Introduction}
% Introducing Transformers
Transformers \cite{vaswani2023attentionneed} are central to modern AI, powering advances in language models, image processing, and beyond. However, their high computational demands lead to substantial energy consumption. Enhancing their efficiency is essential to reduce environmental impact and to keep pace with the exponentially growing size of AI models.
The success of Transformers as state-of-the-art in sequence processing and generation is enabled by their attention mechanism \cite{bahdanau_neural_2016}. To capture dependencies across sequences, the attention mechanism performs dot-products between different projections of multiple sequence elements, known as tokens.
For generative tasks, the best performance is achieved by auto-regressive, decoder-only Transformers \cite{lin_survey_2022}.
At each inference step, the decoder generates a token, which is then appended to the input sequence, forming the input for the subsequent step. 
To avoid recomputing the projections of the previously generated tokens, the so-called \textit{KV-caching} method stores the projections from previous tokens in memory and updates the KV-cache with the new projections \cite{pope_efficiently_2023}. 

In a \ac{GPU}, for each token the entire KV-cache must be transferred from main \ac{HBM} to cache memory (SRAM).
Additionally, the KV-cache is often much larger than the available SRAM memory due to the dimensions of the stored projections and the sequence length \cite{liu_kivi_2023}.
For instance, the entire KV-cache of the model Mistral 7 B \cite{jiang_mistral_2023} requires 8 Gb for a batch size of one, as necessary for inference workloads.
In recent technologies, the energy for data access exceeds the energy required for computations \cite{jouppi_ten_2021}.
Loading the KV-cache for the attention mechanism is thus a major bottleneck, causing increased energy consumption and latency in LLMs \cite{fu2024challengesdeployinglongcontexttransformers}.

% SoTA
To mitigate data transfer overhead, several approaches leverage either near-memory or in-memory computing (IMC)~\cite{10.1145/3400302.3415640, reis_attention--memory_2021, laguna_hardware-software_2022, 10155455, bhattacharjee2024clipformer, wolters_memory_2024, wu_pim_2024}. 
However, the existing solutions do not fill all requirements for efficient hardware attention computation, which are high parallelism for fast inference, high memory density for large implementations, high retention time to avoid required memory refresh, and fast and energy efficient memory writing, as KV-cache is input dependent and must be updated for each generation step. 
KV-cache has been either implemented by DRAM memories \cite{zhou_transpim_2022, wu_pim_2024}, which have limited parallelism  requiring many digital sequential adders, or by SRAM \cite{10155455, liu_hardsea_2024}, which are limited by their volatility and relatively low density \cite{lepri_-memory_2023}. Non-volatile memories can be used for linear layers of Transformers \cite{10.1145/3400302.3415640}, but are too slow, energy expensive, and are not endurant enough for dynamical KV-cache writing \cite{laguna_hardware-software_2022, sebastian2020memory}.

In this work, we propose an in-memory computing hardware architecture based on emerging charge-based memory devices called gain cells \cite{Wang2021, gou20232t1c} to store token projections and compute dot-products for the attention mechanism.
Gain cells store information in a capacitor, with a dedicated read transistor generating current based on the capacitor’s voltage. Unlike DRAM, this enables non-destructive read operations, supporting highly parallel IMC computations. Gain cells have high endurance, fast write speeds, low write energy, and are multi-level. \ac{OSFET}-based gain cells (e.g. \ac{IGZO} or \ac{ITO}) can be integrated in 3D fabrication \cite{lu2024high, shi_counteractive_2023, 10185398, ye_double-gate_2020} and can hold a state for up to multiple seconds without a power supply \cite{shi_counteractive_2023,10185398,ye_double-gate_2020}.

The analog-to-digital conversion required for analog IMC often hinders the advantages this approach offers, as \acp{ADC} are power- and area-intensive \cite{cai2019fully}. To mitigate this issue, charge-based integration is an energy-efficient alternative \cite{wan2022compute, ambrogio2023analog}. Here, we choose to perform the core of the attention mechanism -- two dot-products, scaling, and activation function-- fully in the analog domains, using charge-to-pulse circuits for activation and inter-module communication, combined with pulse counters for final readout. 

Practical applications of \acp{LLM} often rely on pre-trained models to reduce training costs. 
However, our co-optimization approach introduces specific hardware constraints to enhance architectural performance, which leads to a divergence from standard pre-trained models. These constraints include non-ideal multiplication in gain cells and charge-to-pulse circuits, which implement an element-wise ReLU activation function instead of the conventional softmax, which is challenging to realize in analog hardware \cite{electronics10091004}.

To overcome this discrepancy, we introduce an innovative algorithm that adapts a pre-trained language model to our architecture by scaling each layer according to its statistics and hardware characteristics. With our adaptation algorithm, our model achieves accuracy similar to a pre-trained GPT-2 model without having to train the model from scratch.
%Our architecture achieves up to five and two orders of magnitude lower energy consumption and %latency, respectively, in comparison to GPUs for attention computation.
Overall, the contributions of this study are:
\begin{itemize}
    \item An in-memory, mixed analog-digital computing design to store token projections and compute attention dot products with gain cell arrays at high energy-efficiency.
    \item An end-to-end attention mechanism based on analog signals leveraging charge-to-pulse circuits to avoid power- and area-intensive \acp{ADC}.
    \item Quantitative performance analysis of a scalable architecture with area floorplan including analog circuits and digital peripheries.
    \item A software-to-hardware methodology to map pre-trained (ideal) models to non-traditional hardware reaching an accuracy equivalent to GPT-2.
    \item Our architecture achieves up to five and two orders of magnitude lower energy consumption and latency, respectively, in comparison to \acp{GPU}.    
\end{itemize}
After detailing the attention mechanism algorithm, we demonstrate its implementation using gain cells and charge-to-pulse circuits. We then show how our approach maps a pre-trained model to our hardware while maintaining high accuracy on common \ac{NLP} benchmarks. Finally, we evaluate the architecture’s performance in terms of energy consumption, latency, and area footprint.

\begin{figure}[htb!]
    \includegraphics[width=1.\linewidth]{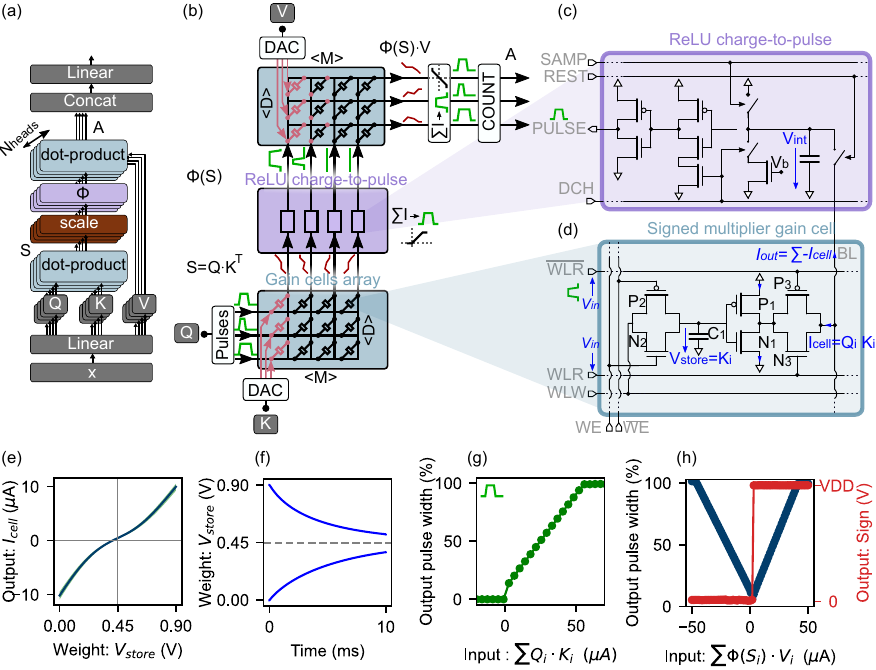}
    \vspace{-0.6cm}
    \centering    
    \caption{(a) Multi-head attention mechanism. (b) Hardware architecture of the attention inference accelerator. (c) Circuit diagram of a gain cell-based signed weight multiplier cell used in the attention computation. (d) Diagram of the ReLU charge-to-pulse circuit. (e) Output current of gain cell based signed weight multiplier %$i(W_{i,j})$
    for different weight voltages $V_{store}$ for $V_{in}=0.9$ V. CMOS process variations are represented by the green shaded area. (f) Silicon CMOS capacitor voltage decay over time due to charge leakage. (g) Behavior of the ReLU charge-to-pulse circuit. (h) Behavior of the signed charge-to-pulse circuit.} 
    \label{fig:concept}
    \vspace{-0.3cm}
\end{figure}

% Now explain what we need for attention (technical)
Fig.~\ref{fig:concept}~(a) shows the attention mechanism algorithm. In auto-regressive Transformers, new token projections called queries (\(Q\)), keys (\(K\)), and values (\(V\)) are created for each inference step from the weights \(W_{Q,K,V} \in \mathbb{R}^{D, d}\) and an input token \(x_i \in \mathbb{R}^{1, D}\) as:
\begin{equation}
    Q_{i}, K_{i}, V_{i} = W_{Q,K,V}x_{i}.
    \label{eq:projections}
\end{equation}
The keys and values \(K_{i} \in \mathbb{R}^{1, d}\) and \(V_{i} \in \mathbb{R}^{1, d}\) are stored as part of the full KV-cache with \(K \in \mathbb{R}^{T, d}\) and \(V \in \mathbb{R}^{T, d}\). 
The query \(Q_{i} \in \mathbb{R}^{1, d}\) is not stored but used for inference as
\begin{equation}   
    \begin{array}{cc}
    S_{i}=Q_{i} \cdot K^T; & \quad A_{i}=\phi \left( \frac{S_{i}}{\sqrt{d}}\right) \cdot V.
    \end{array}
    \label{eq:attention}
\end{equation}
The dot product between the queries and keys produces an attention score matrix \(S_{i} \in \mathbb{R}^{1, T}\). The activation \(\phi\) is often a softmax, but other nonlinear activation functions can yield similar accuracy \cite{katharopoulos_transformers_2020, ma_mega_2023}. The output of the attention mechanism \(A_{i}\) is then obtained by the dot product between the activation \(\phi \left( S_i \right)\) and the values.
In the transformer architecture, multiple attention "heads" are computed in parallel, concatenated and provided to a subsequent linear layer to produce the final multi-head attention result. 

In decoder-based Transformers, causal attention allows the score matrix \(S\) to compare the input token with all previous sequence elements. However, to prevent the physical memory size from scaling with the entire sequence length, we employ a type of attention that is both causal and local: Sliding Window Attention \cite{beltagy_longformer_2020}. In this approach, only a fixed number \(M\) of key and value projections are retained in memory and attention scores for elements older than the last \(M\) are masked. (see Fig.~\ref{fig:sliding_window_attention}~(a)). Although Sliding Window Attention is local at each layer, it can still capture global information in deep networks because the receptive field grows with the number of layers \cite{jiang_mistral_2023}.
\section{Results}

\subsection{End-to-End Analog Hardware Attention}
% merging the gain cell multiplier cell with the end to end attention section by paul
\label{sec:hardware_attention}

In this section, we first give an overview of how our architecture performs operations on analog signals to compute attention. Then, we detail how the different circuits operate. 
Keys $K$ and values $V$ are stored in two gain cell arrays. The query $Q_{i}$ is encoded as \ac{PWM} pulses and is the input of the first array, performing the dot-product \(Q_{i}\cdot K^{T}\). An intermediate charge-to-voltage pulse block integrates the output currents from the first array and outputs \ac{PWM} voltage pulses for the second array, while applying a ReLU activation function (Fig.~\ref{fig:concept}~(c)). The second array, computing $\phi\left(S\right) \cdot V$ is read out using a signed charge-to-voltage pulse block, where the resulting pulse widths are measured by a digital counter.

The proposed gain cell, shown in Fig.~\ref{fig:concept}~(d), contains a write stage for programming the capacitor $C_1$ and a multiplication stage approximating the product between the input and the capacitor voltage. 

The storage capacitor is charged with a multi-level voltage pulse emitted by a \ac{DAC}. The voltage pulse is gated to the designated capacitor by a write-enable (WE) transmission gate. Due to leakage in the storage capacitors, the voltages gradually decay over time. Fig.~\ref{fig:concept} (f) presents the simulated transient response of the storage capacitor voltage $V_{store}$, which corresponds to the cell weight for both extreme values \valunit{0}{V} and \valunit{0.9}{V}. An exponential decay fit of the gain cells leakage reveals that the time constant (i.e., retention time) of our silicon CMOS-based gain cell is $\tau=5$ ms. Note that an \ac{OSFET}-based gain cell can achieve multiple orders of magnitude longer retention times \cite{10185398}.

The multiplication stage generates a current dependent on the stored capacitor voltage (\(V_{store}\)). The input, which is a \ac{PWM} signal, controls the state (closed or open) of the transmission of the multiplication stage. Therefore, the cell generates current only for the input pulse duration. The read path is arranged in a push-pull configuration consisting of two other transistors generating the analog currents. Fig.~\ref{fig:concept}~(e) shows the relation between the output current and the voltage stored in the capacitor $V_{store}$ with one cell simulated through SPICE simulations including Monte-Carlo sampling for variability analysis. Since multiple cells are connected to a shared Bit Line (BL), their currents are summed according to Kirchhoff's law.

In each inference step, both arrays are updated with one column from the key and value matrices, as we will show in more detail in Section~\ref{sec:sliding_window}. The \( M \) columns of each array represent the \(K\) and \(V\) of the previous \( M \) tokens, while the rows correspond to the \(d\) distinct embedding elements.

Since the gain cell array inputs are temporally encoded, their output currents are also in the temporal domain and thus need to be integrated to retrieve the correct dot-product results. The signal between the two arrays is converted by the charge-to-pulse circuit depicted in Fig.~\ref{fig:concept}~(c) that integrates the currents and emits a \ac{PWM} voltage pulse of variable width depending on the accumulated charge, similarly as in \cite{10114059} (see~Fig.~\ref{fig:concept}~(g)). The pulse is emitted only if the charge is positive, following the ReLU function. Then, the pulse width increases linearly with the charge, until it saturates to the maximum pulse width when the circuit reaches a threshold \(S_{sat}\). More details are available in the supplementary material% section~\ref{supp:ReLU}
.

The pulses representing $\phi\left(S\right)\in \mathbb{R}^{M}$ are fed as inputs to the second gain cell array to perform the dot product $\phi\left(S\right) \cdot V$. A different type of charge-to-pulse circuit integrates the output currents of the second array. Unlike the first one, this signed charge-to-pulse circuit is capable of generating pulses for both positive and negative input charges, while a D-Flip Flop stores the result's sign. The behavior of this circuit for different inputs is highlighted in Fig.~\ref{fig:concept}~(h). A 16-level digital counter measures the generated pulse widths and multiplies the result by the retrieved sign bit, resulting in a total precision of 32 levels.

\subsection{Analog Hardware Sliding Window Attention Data-flow}
\label{sec:sliding_window}
\begin{figure}[htb!]
    \includegraphics[width=1.\linewidth]{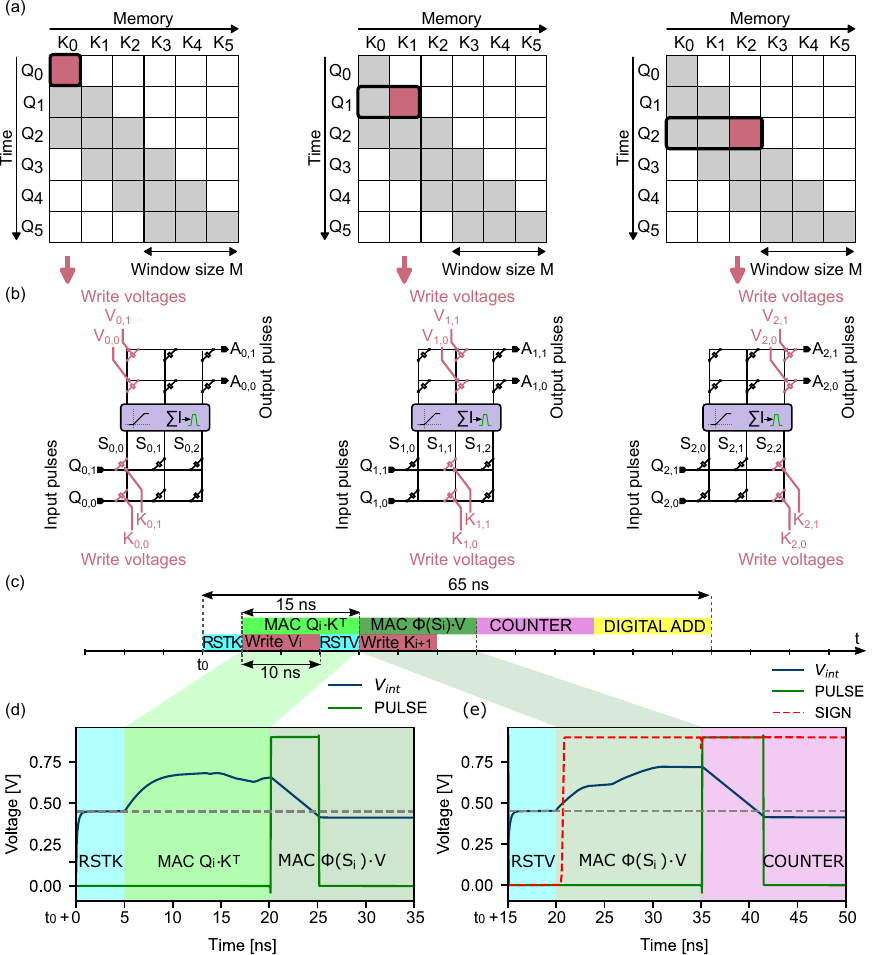}
    \vspace{-0.6cm}
    \centering    
    \caption{\raggedright (a) Three inference steps of a dot-product between \(Q\) and \(K\) in Sliding Window Attention. The grey boxes represent tokens that are attended to, and the blank boxes the masked tokens. (b) Equivalent gain cell array implementations for an entire attention head. A new column of \(K\) and \(V\) is written  before each inference step. (c) Proposed pipeline, highlighting parallel operations of writing new $K-V$ pairs and performing the MAC operations. (d) Transient Simulation of the $\Phi(Q \cdot K^T) $ MAC operation including temporal location. (e) Transient Simulation results of the $\Phi(S) \cdot V$ MAC operation including the pulse and sign signal for the counter within the pipeline.}
    \label{fig:sliding_window_attention}
    \vspace{-0.3cm}
\end{figure}

Having described how inference is performed for one token, we now describe how the architecture processes multiple tokens sequentially.
In Sliding Window Attention, the input query is multiplied only with the \(M\) most recent keys and values, corresponding to the window size $M$ (see~Fig.~\ref{fig:sliding_window_attention}~(a)). At each time step, the keys and values must be updated with the most recent token and the oldest one must be forgotten. All other projections remain stationary until they are updated after $M$ cycles. In our implementation, we write the array that encodes the keys and values at inference time in a column-wise manner (see~Fig.~\ref{fig:sliding_window_attention}~(b)). 

Before writing the gain cells, the capacitors are reset in a \valunit{5}{ns} discharge phase.
At time step \(t=0\), \(d\) cells of the first column using the vectors \(K_{0}\) and \(V_{0}\) are written by charging the corresponding capacitors with \valunit{10}{ns} multi-level pulses generated by \valunit{3}{bit} \ac{DAC}s. Inference is then performed with inputs \(Q_{0}\), encoded by pulses with durations between \valunit{0}{ns} and \(T_{max}\) = \valunit{15}{ns} generated by \valunit{4}{bit} ($16$~levels) voltage pulse generators with \valunit{1}{GHz} clock speed. For the next time step \(t=1\), we write the cells of the second column using the different \(K_{1}\) and \(V_{1}\) and repeat the attention computation with \(Q_{1}\). After \(M\) time steps, when all the columns of the array have been written, we overwrite the first column of the array, thus  forgetting the oldest stored key. The succession of the different write and read cycles implements a sequential Sliding Window Attention. The read phase of one \(K\) array can therefore be pipe-lined with the write phase of the next \(V\) array as described in Fig.~\ref{fig:sliding_window_attention}~(c).
\subsection{Full Attention Head Hardware Implementation}
\begin{figure}[htb!]
    \includegraphics[width=1.\linewidth]{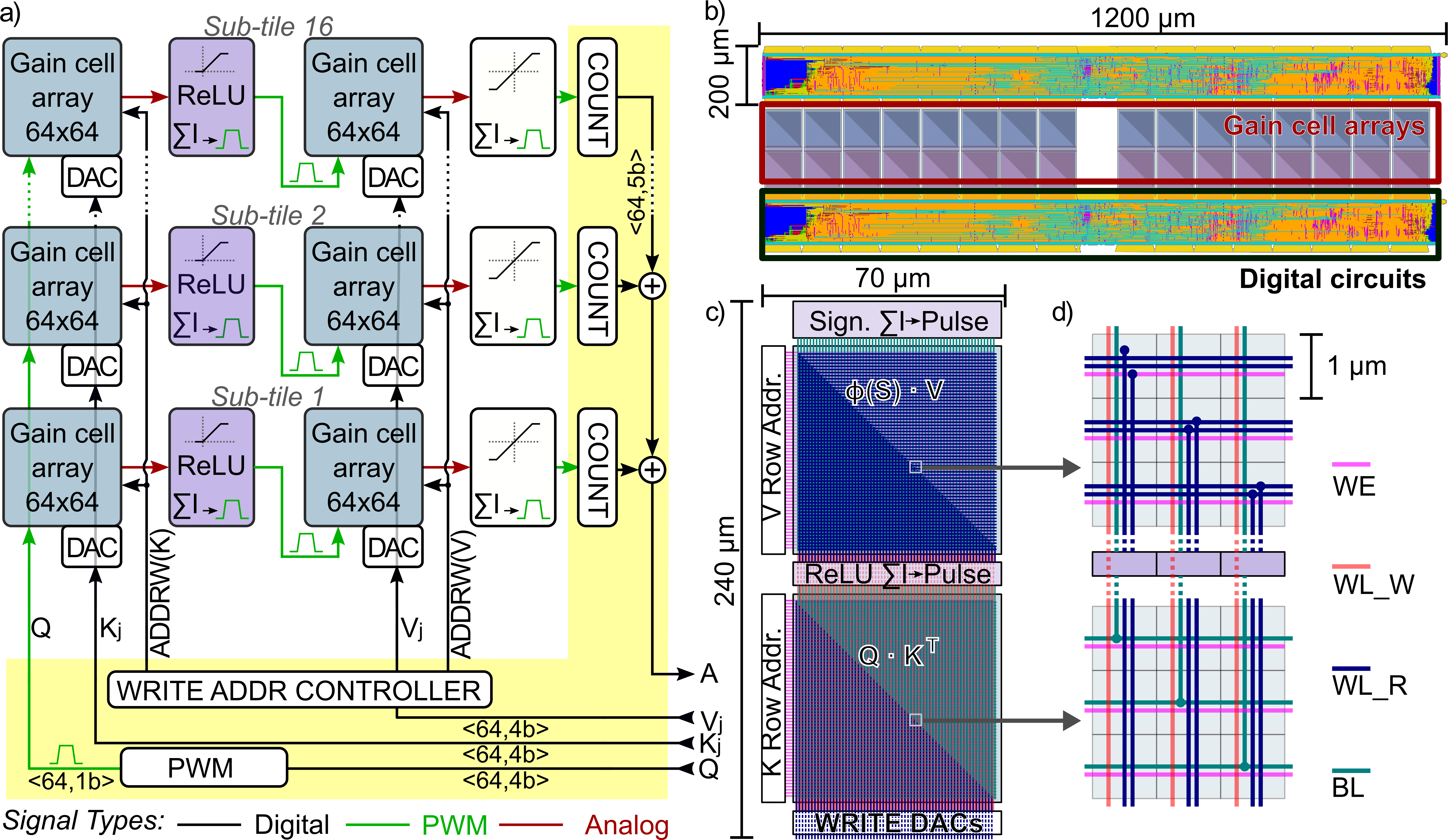}
    \vspace{-0.6cm}
    \centering    
    \caption{\raggedright (a) Proposed hardware architecture for a single attention head, featuring the tiling of the attention head into multiple sub-tiles. The digital peripheral control is highlighted in yellow. (b) Floorplan of the architecture for one attention layer using \ac{OSFET} technology assumptions. (c) Floorplan of one sub-tile. (d) Routing of one sub-tile highlighting the vertical propagation of signals by diagonal wire tapping.
}
    \label{fig:architecture}
    \vspace{-0.3cm}
\end{figure}
% Explain tiling but also a sentence to say we do multiple heads
IR drop, caused by resistive losses in interconnects, results in reduced accuracy in large-scale analog crossbar arrays~\cite{10423426}. To mitigate IR drop issues, we limit the size of our gain cell arrays to 64 $\times$ 64. However, most \ac{NLP} applications require larger either larger window dimension $M$ (columns) or larger embedding dimension \(d\) (rows). To accommodate larger dimensions, we perform inference across multiple sub-tiles, as shown in Fig.~\ref{fig:architecture}~(a).

In this paper, we implement a GPT-2 model with an embedding dimension \(d\) = 64 and a sliding window size \(M\) = 1024. Therefore, the entire KV-cache of the window size \(M\) is divided into 16 sub-tiles, each having its charge-to-pulse blocks and storing a fraction of the $K$ and $V$ in two 64 $\times$ 64 arrays. A write address controller keeps track of the current write index.
All tiles receive the same input \(Q \)  generated by the digital block in parallel, are measured by pulse counters, and summed by 64 digital adders, each with 16 inputs (see Fig.~\ref{fig:architecture}~(b,~c)). 

\subsection{Pre-trained Model Hardware-Aware Mapping and Fine-Tuning}
\label{sec:training_algo}
\begin{figure}[htb!]
    \includegraphics[width=1.\linewidth]{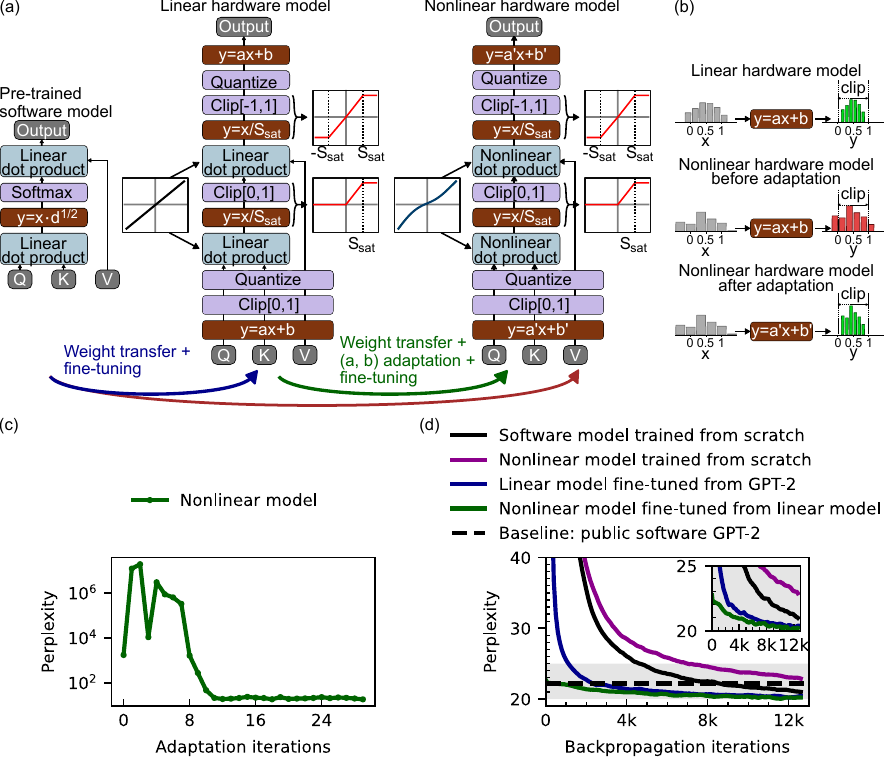}
    \vspace{-0.6cm}
    \centering    
    \caption{(a) Pre-trained model mapping. From a software pre-trained model, we fine-tune an intermediate model that integrates all hardware constraints except dot-product nonlinearity. Then, we use a custom adaptation algorithm to map the intermediate model to the gain cell's nonlinearity. Finally, we fine-tune the model nonlinear model. (b) Sketch of the adaptation algorithm for scaling factors. Scaling factors re-scales the input before clipping and quantization. The nonlinear model leads to different statistics (red histogram) from the linear model (green histogram). The adaptation algorithm modifies the scaling factors to match the statistics of the nonlinear model to the the statistics of the linear one. (c) Evolution of perplexity (lower the better) during the adaptation algorithm.
    (d) Training curves for the different models. The software model is GPT-2, the nonlinear model is the model with the proposed hardware attention, and the linear model is the hardware attention with ideal linear gain cells.} 
    \label{fig:algo}
    \vspace{-0.3cm}
\end{figure}

Using weights from pre-trained models is challenging because our attention mechanism differs from the conventional ones (see~Fig.~\ref{fig:algo}~(a)). The main differences are:
\begin{itemize}
    \item ReLU activation used instead of softmax (see~Fig.~\ref{fig:concept}~(b)).
    \item Sliding Window Attention is implemented instead of Causal Attention (see~Fig.~\ref{fig:sliding_window_attention}~(a)).
    \item Input, stored projections, and output are quantized in four, three, and five bits respectively by digital \ac{PWM}s, \ac{DAC}s, and pulse counters (see~Fig.~\ref{fig:concept}~(b)).
    \item Gain cell arrays are split into sub-tiles before final result summation (see~Fig.~\ref{fig:architecture}~(a)).
    \item The relation between gain cell input and stored voltages is nonlinear (see~Fig.~\ref{fig:concept}~(e)). 
    \item Capacitor leakage causes stored value decay (see~Fig.~\ref{fig:concept}~(f)).
\end{itemize}
The implementation of these hardware constraints in our simulations is explained in Methods~\ref{sec:method-hardware-neural-sim}. As the non-linear relation between input voltage and stored voltage in gain cells is described by a third-order polynomial function, this significantly increases the computational complexity and memory requirements to train our gain cell-based model. Therefore, to adapt the pre-trained public GPT-2 model to our hardware constraints, we first fine-tune it using an intermediate model. The intermediate model employs ideal linear dot-products, but integrates all the other mentioned hardware constraints. 
The model is trained on predicting the next words of the open-source text collection OpenWebText \cite{Gokaslan2019OpenWeb}, and the metric used for evaluation is perplexity, which measures the uncertainty of the prediction (lower perplexity means higher prediction confidence). In \ref{fig:algo}~(d), we see that our linear intermediate model (blue curve) achieves results equivalent to a public GPT-2 model in less than 3,000 iterations, whereas it takes more than 13,000 iterations for the model trained from scratch (magenta curve). This result shows that performing weight transfer is efficient even though the two models are different (in particular, ReLU activation instead of softmax).

After fine-tuning the intermediate linear model, we transfer the weights to the final hardware model including the gain cell's nonlinearity. This mapping is non-trivial, as all the layers have different statistics, making it difficult to apply a single fit to capture the gain cells' nonlinearity. To circumvent this issue, we introduce scaling operations.
\begin{equation}
    y=ax+b
    \label{eq:scaling}
\end{equation}
with distinct scalars \(a\) and \(b\) for each of the \(Q\), \(K\), and \(V\) projections, as well as for the output of the attention, with separate factors applied across different attention heads and layers.

To choose the scaling parameters \(a\) and \(b\), we develop an algorithm inspired by \cite{mishkin_all_2016}.
Given a set of input samples, we use an iterative loop that updates the scaling parameters so that the output of the scaling function of the nonlinear model matches the statistics of the linear model (as sketched in Fig.~\ref{fig:algo}~(b)). First, we measure the standard deviation $\sigma_{L}$ and the mean $\mu_{L}$ of the output of every scaling stage (see equation~(\ref{eq:scaling})) of the linear model on a large set of samples. Then, at each iteration, we measure the standard deviation $\sigma_{NL}$ and the mean $\mu_{NL}$ for the scaling stage of the nonlinear model. For each iteration, the scaling parameters are updated as
\begin{equation}
    \begin{split}
        a &\leftarrow a \frac{\sigma_{L}}{\sigma_{NL}} \\
        b &\leftarrow b + \left(\mu_{L} - \mu_{NL} \right)
    \end{split} \text{.}
\end{equation}
In Fig.~\ref{fig:algo}~(c), we show how the perplexity of the nonlinear gain cell model is reduced from 1757 to 21 during this adaption stage. In supplementary material %\ref{supp:non_linear_adapt}
 we show that this adaptation algorithm can generalize to other multiplication nonlinearities.
After the adaptation algorithm, we can fine-tune the nonlinear model using backpropagation (see~Fig.~\ref{fig:algo}~(d), green curve) to further improve the results. The entire process is described in Fig.~\ref{fig:algo}~(a).

\subsection{Downstream Task Benchmarks}
To demonstrate the efficacy of the proposed hardware attention and mapping approach, we evaluate two software baselines and three hardware models on standard language model bench-marking tasks (see~Table \ref{tab:lm_eval_results}), which are explained in the Methods \ref{sec:method-dowstreasm-tasks}.

Our nonlinear model which is both adapted from the linear model and fine-tuned achieves a precision very close or equal to the public software model, and equivalent or higher than the software model trained from scratch with the same training procedure. These results suggest no apparent limitations for \ac{NLP} tasks using our hardware attention. We also tested our nonlinear model without nonlinearity fine-tuning, observing equivalent performance across most datasets, except for LAMBADA and WikiText-2. This is promising, as it demonstrates that our approach yields competitive results without requiring precise device characteristics during backpropagation.
\begin{table}[t]
    \centering
    \resizebox{\textwidth}{!}{%
\begin{tabular}{l c c c c c c c c } 
\toprule 
  & ARC-E  & ARC-C  & WinoGrande  & HellaSwag  & LAMBADA  & LAMBADA  & PIQA  & WikiText-2 \\ 
 & acc $\uparrow$  & acc $\uparrow$  & acc $\uparrow$  & acc $\uparrow$  & ppl $\downarrow$  & acc $\uparrow$  & acc $\uparrow$  & ppl $\downarrow$ \\ 
\midrule 
 Public software model & 43.81  & 22.70  & 51.62  & 31.14  & 40.06  & 32.56  & 62.89  & 37.37 \\  \midrule 
 Software model trained from scratch & 42.34  & 23.46  & 50.20  & 29.73  & 53.43  & 29.69  & 61.48  & 41.25 \\ \midrule 
 Linear hardware model & 42.80  & 23.46  & 52.41  & 30.31  & 58.66  & 25.54  & 61.21  & 39.79 \\ \midrule 
 \makecell[l]{\textbf{Nonlinear hardware model} \\ \textbf{with adaptation}} & \textbf{42.09}  & \textbf{22.87}  & \textbf{50.51}  & \textbf{30.10}  & \textbf{89.08}  & \textbf{19.48}  & \textbf{61.53}  & \textbf{42.34} \\ \midrule 
 \makecell[l]{\textbf{Nonlinear hardware model} \\ \textbf{with adaptation and fine-tuning}} & \textbf{43.94}  & \textbf{22.78}  & \textbf{51.14}  & \textbf{30.18}  & \textbf{49.49}  & \textbf{27.87}  & \textbf{62.62}  & \textbf{39.97} \\ 
 \bottomrule 
\end{tabular} 
    } % end resizebox
    \caption{Downstream task results. The metrics are the percentage of accurate word predictions (acc), and the perplexity (ppl), a measure of prediction  uncertainty.
    }
    \label{tab:lm_eval_results}
\end{table}
\subsection{Circuit Computing accuracy}
The accuracy of our circuits for attention computation is highlighted in Fig.~\ref{fig:energy_latency}~(a, b). For each of the two dot dot-products, we simulate one 64$\times$64 array and the corresponding 64 charge-to-pulse circuits. The results of the first dot-product, which are shown in Fig.~\ref{fig:energy_latency}~(a), are fed as input to the second dot-product and are shown in Fig.~\ref{fig:energy_latency}~(b). For each plot, we compare the SPICE simulations with the model used for neural network simulations.

\subsection{Energy Consumption and Latency}
\begin{figure}[htb!]
    \centering
    \includegraphics[width=1.\linewidth]{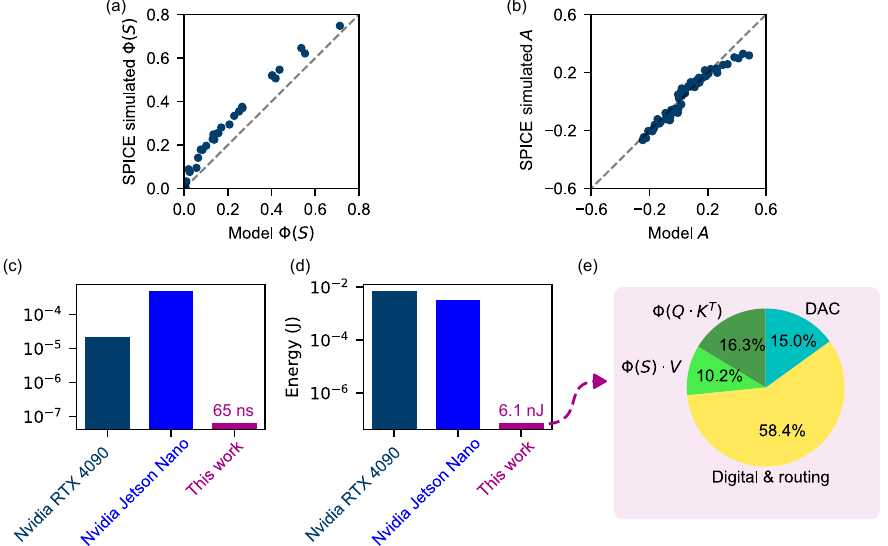}
    \caption{(a) Comparison of expected results model versus Spice simulation results for the $\Phi(Q \cdot K^T)$ operation. (b) Comparison of PyTorch model versus SPICE simulation results for the $\Phi(S) \cdot V$ operation. (c) Latency of the attention mechanism for one processed token and (d) energy consumption for a twelve head attention mechanism implemented by a consumer GPU, an embedded application-specific GPU, and our hardware architecture. (e) Energy consumption ratio for the different modules of our hardware architecture, including analog and digital signals.}
    \label{fig:energy_latency}
\end{figure}
The circuit's operational speed and timing, on which the energy assumptions are based are shown in Fig.~\ref{fig:sliding_window_attention}~(d). The total latency of attention can be estimated to \valunit{65}{ns}. 

The gain cell arrays and charge-to-pulse circuits consume \valunit{1120}{pJ} per token computation for the first dot-product, and \valunit{700}{pJ} for the second dot-product. The lower energy consumption in the second dot-product arrays is attributed to the significantly sparser activation of its input $\Phi(S)$, leading to less currents in the second gain cell array. The digital control and routing block consumes a total power of \valunit{113.7}{mW}, or \valunit{4}{nJ} per token, while the \ac{DAC}s require \valunit{330}{pJ}. Overall, we can estimate the power consumption of processing one token for one attention head to \valunit{6.1}{nJ}. A pie chart of the power composition attributed to each unit is shown in Fig.~\ref{fig:energy_latency}~(e).

The energy and latency of our architecture, compared with two different GPUs, are shown in Fig.~\ref{fig:energy_latency}~(c,d). Focusing on the attention mechanism alone, our architecture can lead to a speedup of $\times  7,000$ compared to Nvidia Jetson Nano and $\times  300$ compared to Nvidia RTX 4090, as well as an energy reduction of $\times  40,000$ compared to Jetson Nano and $\times  90,000$ compared to RTX 4090.

\subsection{Area and Floorplan}
To demonstrate the scalability and potential area efficiency of our architecture, we base our floorplan on emerging \ac{OSFET}-based gain cells, which have high density potential as they can be integrated in 3D, and do not require large capacitors since \ac{OSFET} write transistors have very low leakage currents~\cite{liu_design_2024, subhechha_demonstration_2023}. 
Based on our assumptions, described in Methods~\ref{sec:method-area}, for the worst-case scenario, the area of the proposed gain cell is \valunit{1}{um^2}. Fig.~\ref{fig:architecture}~(c), shows the floorplan of a single tile, which includes 64 shared \acp{DAC} for writing the weights, two-row address decoders, and charge-to-pulse circuitry. The total area of one head, shown in the floorplan in Fig.~\ref{fig:architecture}~(b), is \valunit{0.5}{mm^2}. Multiple attention heads can be operated by multiple tiles in parallel on the chip, and can be stacked in 3D with shared peripheral and digital logic.

\section{Discussion}
In this work, we proposed a novel analog in-memory computing architecture to improve the energy consumption and latency of the attention computations at the core of generative AI models. 

Our design leverages capacitor-based gain cells, offering an efficient solution for both memory storage and computation, significantly improving energy efficiency and speed. 
To avoid power-intensive \acp{ADC}, we perform the attention computation in the analog domain, using charge-to-pulse circuits to transmit analog signals between computation stages. This approach introduces non-ideal operations compared to digital attention computations, but with substantial efficiency gains. But another significant contribution is a hardware-aware training methodology compensating for the circuit non-idealities. Nonetheless, future circuit optimizations could further reduce any discrepancies. 

Our neural network simulations confirm that an \ac{LLM} implemented with our hardware attention achieves results comparable to software-based networks, even on complex NLP tasks. While our study uses device-level simulations to evaluate design performance, our adaptation algorithm demonstrates potential for measured device implementations, as it allows most of the training process to proceed without requiring precise device-specific models of nonlinear behavior, making the approach generically applicable and computationally efficient.

Our architecture can benefit from \ac{OSFET} transistors that enable dense 3D integration~\cite{liu_design_2024, subhechha_demonstration_2023}, offering potential for very compact implementations of large networks. Latency is reduced by up to two orders of magnitude, and energy consumption by up to five orders for attention computations alone compared to \acp{GPU}. While we focus on the attention mechanism, a major bottleneck in generative Transformers' inference, significant reductions in overall energy consumption require optimizing all components. In the future, our hardware attention mechanism can be integrated with other IMC techniques to implement low-power linear layers.

In conclusion, this work demonstrates hardware-algorithm co-optimization achieving low latency and energy consumption while maintaining high model accuracy. Additionally, it highlights the promise of in-memory computing with volatile, low-power memory for attention-based neural networks, marking an important step toward ultra-fast, energy-efficient generative AI.
%TC:ignore
\section{Methods}
\subsection{Hardware-based Neural Network Simulations}
\label{sec:method-hardware-neural-sim}
We implement the Sliding Window Attention by masking the elements of \(S\) outside the sliding window (blank spaces in the example Fig.~\ref{fig:concept}). The ReLU charge-to-pulse circuit is modeled by the equation
\begin{equation}
        \phi(S) = \begin{cases} 
          T_{max} & \text{if } S \geq S_{sat} \\
\frac{T_{max}}{S_{sat}}S & \text{if } 0 < S < S_{sat} \\
          0 & \text{if } S \leq 0
        \end{cases}
        \text{,}
    \label{eq:analogReLU}
\end{equation}
where \(T_{max}\) = \valunit{15}{ns} as for the input pulse generators. The input query \(Q\) are quantized in 16 levels between 0 and 1, the stored \(K\) and \(V\) projections are quantized in 8 levels between 0 and 0.9, and the output of the second dot-product are quantized in 32 levels between -1 and 1. The quantized models (linear intermediate hardware model and non-linear hardware model) are trained with Quantization Aware Training \cite{jacob_quantization_2017}: quantization is done only in the forward pass and the backward pass is done in full precision.

%The non-linear model of the gain cells is performed through a third-order polynomial fit of their I-V characteristics (see~Fig.~\ref{fig:concept}~(e)). 
% For the linear intermediate model the equation 
% \begin{equation}
%     \begin{split}
%         S=Q\cdot\left(K^{T}-K_{offset}\right)V_{in}C \\
%         A=\phi\left(S\right) \cdot \left(V^{T}-V_{offset}\right)V_{in}C,
%     \end{split}
%     \label{eq:linearGainCell}
% \end{equation}
% is used as a linear approximation of the gain cells for respectively the first and the second dot products, with \(S\) as the output, $Q$ and $\phi\left(S\right)$ the input pulse width, \(K\) and \(V\) the stored voltage, the constant $V_{in}=$ 0.9 V is the input voltage of the cell, the constant $y_{offset}=$ 0.45 V corresponds to $V_{dd}/2$, and $C$ is a linear fit parameter from Fig.~\ref{fig:concept}~(e).
For the non-linear model of the gain cell, the third-order polynomials
\begin{equation}
    \begin{split}
        S = \sum_i^3\sum_j^{3-i} Q \cdot \left(K^{T}-K_{offset}\right)^{i}V_{in}^{j}C_{i,j}\\
        A = \sum_i^3\sum_j^{3-i} \phi\left(S\right) \cdot \left(V-V_{offset}\right)^{i}V_{in}^{j}C_{i,j}
    \end{split}
    \label{eq:nonlinearGainCell}
\end{equation}
are used with \(S\) and \(A\) as the outputs, $Q$ and $\phi\left(S\right)$ the input pulse width, \(K\) and \(V\) the stored voltage, the constant $V_{in}=$ 0.9 V is the input voltage of the cell, the constant $y_{offset}=$ 0.45 V corresponds to $V_{dd}/2$, and $C_{i,j}$ as fit parameters from the curve Fig.~\ref{fig:concept}~(e). To speed-up computation during training, we compute all the tokens in parallel with $Q \in \mathbb{R}^{T, D}$, $K^{T} \in \mathbb{R}^{D, T}$, $V \in \mathbb{R}^{T, D}$, and $\phi\left(S\right) \in \mathbb{R}^{T, T}$ (the batch dimension and the head dimension are omitted for simplicity).

The capacitor leakage leads to an exponential decay in the stored value. After discretization, the exponential decay is formulated as 
\begin{equation}
    \begin{array}{cc}
    y_t=y_{t-1} e^{-\frac{\Delta_{t}}{\tau}}; & \quad \Delta_{t}=L\delta_{t},
    \end{array}
    \label{eq:discrete_decay}
\end{equation}
where $\tau$ is the time constant of the capacitors, $\Delta_{t}$ is the time elapses between two inference steps $\delta_{t}$ is the latency caused by each neural network layer, and $L$ is the number of layers. To model the decay of all capacitors at all time steps in parallel, we introduce a decay mask $\alpha \in \mathbb{R}^{T, T}$ defined as
\begin{equation}
    \begin{array}{cc}
        \alpha = e^{-\frac{\Delta_{t}}{\tau} m_{t,t'}}; & \quad m_{t,t'} = max\left(0, i-j\right).
    \end{array}
    \label{eq:decay_mask}
\end{equation}
To optimize computation, the decay mask is directly integrated in the dot product computation as 
\begin{equation}
    \begin{split}
        S = \sum_i^3\sum_j^{3-i} \left(Q \cdot \left(K^{T}-K_{offset}\right)^{i}V_{in}^{j}C_{i,j}\right) \alpha^{i}\\
        A = \sum_i^3\sum_j^{3-i} \left(\phi\left(S\right) \alpha^{i} \right) \cdot \left(V-V_{offset}\right)^{i}V_{in}^{j}C_{i,j}
    \end{split}
    \label{eq:nonlinearGainCell}
\end{equation}

In our simulation, we chose a time constant \(\tau =\) 5 ms to be consistent with the data from Fig.~\ref{fig:concept}~(h). We chose \(\delta_{t} =\) 65 ns to be equal to the latency of our full hardware attention mechanism (see Fig.~\ref{fig:sliding_window_attention}~(c)). Our decay factor is therefore $\frac{\Delta_{t}}{\tau}=\frac{12\times65\times10^{-9}}{5\times10^{-3}}\simeq 1.6\times10^{-4}$. In a full Transformer implementation, the latency per layer \(\delta_{t} =\) will be higher than 65 ns as it will also include latency from other modules. However, time constant $\tau$ of three orders of magnitude larger were reported in \ac{OSFET}-based gain cell memories \cite{Wang2021, 10185398}, and therefore we conclude that the choice of decay factor of $1.6\times10^{-4}$ is very conservative. It is noteworthy that the decay of stored keys and values may not necessarily hinder network performance: several approaches in Deep Learning leverage exponential decay masks to enhance memory structure \cite{press_train_2022, ma_mega_2023, wang_state-space_2023}. 

For the adaptation, the algorithm was repeated until the mean and standard deviation of the output of the scaling functions of the nonlinear model matches the mean and standard deviation of the linear model within a tolerance ratio: $\left|\sigma_{NL}-\sigma_{L}\right|<0.0001$ and $\left|\mu_{NL}-\mu_{L}\right|<0.0001$.

\subsection{Hardware-based Neural Network Training}
To evaluate our training algorithm and the inference accuracy of our architecture, we implement the analog gain cell-based attention mechanism on the GPT-2 architecture \cite{radford2019language}. GPT-2 is a Transformer neural network with 124 million parameters, 12 layers, an attention mechanism input dimension of 768, 12 heads per attention block, and a head dimension of 64. We used the open-source text collection OpenWebText \cite{Gokaslan2019OpenWeb} split between training and testing samples, and the pre-trained GPT-2 tokenizer to encode the plain text into tokens (vectors of size 50,304 each). Each training iteration had a batch size of 1,920, with sequences of length 1024 per sample. We selected a sliding window size of 1024, which matches the number of gain cell rows in the memory. Since the sequence length also equals 1024, each gain cell is written only once per sequence, eliminating the need to overwrite cells during one sliding window iteration. For a larger sequence length, the gain cells would be overwritten, as described in Section \ref{sec:sliding_window}. To train the network, the next token in the sequence is predicted for each input token. Thus, the target sequences are the input sequences shifted by one token. The cost function used was cross-entropy, calculated between the predicted sequence and the target sequence. We used backpropagation with the AdamW optimizer \cite{loshchilov_decoupled_2019}, with a learning rate of $6\times10^{-4}$ and a weight decay of 0.1. The results of each evaluation are averaged over 4,000 samples.

\subsection{Downstream Tasks Set-up}
\label{sec:method-dowstreasm-tasks}
The datasets cover various types of problems. Our benchmarking setup is inspired by \cite{gu2024mambalineartimesequencemodeling} and \cite{beck2024xlstmextendedlongshortterm} in terms of evaluated tasks and metrics.
ARC-Easy and ARC-Challenge \cite{Clark2018ThinkYH} focus on question answering, with ARC-Easy containing straightforward questions and ARC-Challenge featuring more difficult ones. WinoGrande \cite{sakaguchi2019winogrande} evaluates commonsense reasoning and co-reference resolution by presenting minimal pairs to resolve ambiguities. HellaSwag \cite{zellers2019hellaswag} tests commonsense inference, requiring models to predict the most plausible continuation of a given context. LAMBADA \cite{paperno2016lambadadatasetwordprediction} evaluates models' text understanding through a word prediction task that requires comprehension of broader discourse, not just local context. PIQA \cite{Bisk2020} assesses physical commonsense reasoning, testing a model’s understanding of physical scenarios. Wikitext-2 \cite{merity_pointer_2016} is a general text corpus derived from Wikipedia articles to assess long-term dependencies processing, text prediction and generation capabilities.% , while OpenBookQA \cite{OpenBookQA2018} measures the ability to combine scientific facts with commonsense reasoning to answer questions.
For fair comparisons, except for software public GPT-2, all the models were evaluated after the same number of training iterations. The linear hardware model was trained on 13,000 iterations, the nonlinear hardware model was mapped from the 13,000 iterations linear model using the adaptation algorithm but without fine-tuning, and the nonlinear hardware model with adaptation and fine-tuning was adapted from a linear model trained on 3,000 iterations, and then fine-tuned on 10,000 iterations.

\subsection{Hardware SPICE simulations}
To assess circuit performance accuracy, energy consumption and speed, we conducted SPICE array simulations using the TSMC \valunit{28}{nm} PDK within the Cadence Virtuoso environment. All simulations are based on a 64$\times$64 array, corresponding to the tile size in our architecture (see Fig.~\ref{fig:architecture}~(a)). In these simulations, a parasitic wire capacitance of \valunit{0.8}{fF} and a series resistance of \valunit{2}{\Omega} per array element are included. Both arrays, one performing $\Phi (Q \cdot K^T)$ and the other performing $\Phi (S) \cdot V$, are simulated separately, but always in combination with their specific charge-to-pulse circuitry readout circuitry.

\subsection{GPU Attention Latency and Energy Consumption Measurements}
To measure the latency and energy on Nvidia RTX 4090 and Jetson Nano, which are respectively a consumer \ac{GPU} and an embedded application \ac{GPU}, we perform ten runs of 1024 steps of auto-regressive token generation with twelve attention heads. For a fair comparison, the linear projections are not implemented in this experiment since they are also not implemented by our hardware architecture, and the static power measured before inference is subtracted from the power measured during inference. For each run, we measure the latency and the power using the Nvidia-SMI python API, and average them.

\subsection{Area Estimation} 
\label{sec:method-area}
Our floorplan is based on \ac{ITO} gain cells, an emerging \ac{OSFET} technology that has enabled low area gain cell designs~\cite{liu_design_2024}. 
A two transistor \ac{ITO} gain cell occupies an area of \valunit{0.14}{um^2} ($\approx 370\,nm \times 370\,nm$)~\cite{liu_design_2024}, allowing for denser memories than CMOS-based gain cells.
%The monolithic 3D stacking capability of these devices further improves the memory density and enables the increase of memory capacity required for LLMs. 
%A few studies~\cite{subhechha_demonstration_2023} have demonstrated the multi-level capabilities of gain cells using \ac{IGZO} devices.
Based on the area results presented in these works~\cite{liu_design_2024, subhechha_demonstration_2023}, we estimate the worst-case area of the proposed six-transistor cell to be \valunit{1}{um^2}, leading to a $19\times$ area reduction compared to gain cells based on CMOS write transistors (our CMOS-based gain cell layout is presented in Supplementary material%\ref{supp:CMOS_layout})
.
The total area of one attention head is derived from this single cell area estimation, as well as the charge-to-pulse circuit layout and the total floorplan incorporating the 16 sub-tiles and digital circuits, providing a precise representation of the space requirements.
This structure is designed to be repetitive (vertical dimension in Fig.~\ref{fig:architecture}~(c)), allowing multiple attention heads to be efficiently integrated on a single chip. 
Each attention head receives inputs from the lower digital block, while its outputs are processed by the upper digital block. To facilitate the connection of the bitline outputs of one array (i.e. vertical metal lines) to the wordline input of the next array (i.e. horizontal metal line), we employ wiretapping, as highlighted in Fig.~\ref{fig:architecture}~(d).

\backmatter

\bmhead{Supplementary information}

An attached supplementary document provides additional details and results from complementary experiments.
The Python scripts used for these experiments are available at https://github.com/NathanLeroux-git/GainCellAttention/.

\bmhead{Acknowledgements}
This work was supported in part by the Federal Ministry of Education and Research (BMBF, Germany) in the project
NEUROTEC II (Project number: 16ME0398K). This work is based on the Jülich Aachen Research Alliance (JARA-FIT) at Forschungszentrum Jülich GmbH, Jülich, Germany. \\
The authors gratefully acknowledge the Gauss Centre for Supercomputing e.V. (www.gauss-centre.eu) for funding this project by providing computing time through the John von Neumann Institute for Computing (NIC) on the GCS Supercomputer JUWELS at J\"ulich Supercomputing Centre (JSC).

\noindent

%\bibliography{references.bib}

%%% references

%%% end references

%\include{supp_include.tex}

%TC:endignore
\end{document}

%TC:ignore
\begin{acronym}
    \acro{MAC}{Multipy-Accumulate}
    \acro{IMC}{In-Memory-Computing}
    \acro{LLM}{Large Language Model}
    \acro{DRAM}{dynamic Random Access Memory}
    \acro{ASIC}{Application-Specific Integrated Circuit}
    \acro{FPGA}{Field-Programmable Gate Array}
    \acro{PWM}{Pulse-Width Modulation}
    \acro{DAC}{Digital to Analog Converter}
    \acrodefplural{DAC}[DACs]{Digital to Analog Converters}
    \acro{ADC}{Analog to Digital Converter}
    \acrodefplural{ADC}[ADCs]{Analog to Digital Converters}
    \acro{IGZO}{Indium Gallium Zinc Oxide}
    \acro{NLP}{Natural Language Processing}
    \acro{HBM}{High Bandwidth Memory}
    \acro{GPU}{Graphical Processing Units}
    \acro{OSFET}{Oxide Semiconductor Field Effect Transistor}

\end{acronym}

\title[Article Title]{Supplementary Material: Analog In-Memory Computing Attention Mechanism for Fast and Energy-Efficient Large Language Models}

\maketitle

\section{CMOS layout}
\label{supp:CMOS_layout}
\begin{figure}
    \centering
    \includegraphics[width=1\linewidth]{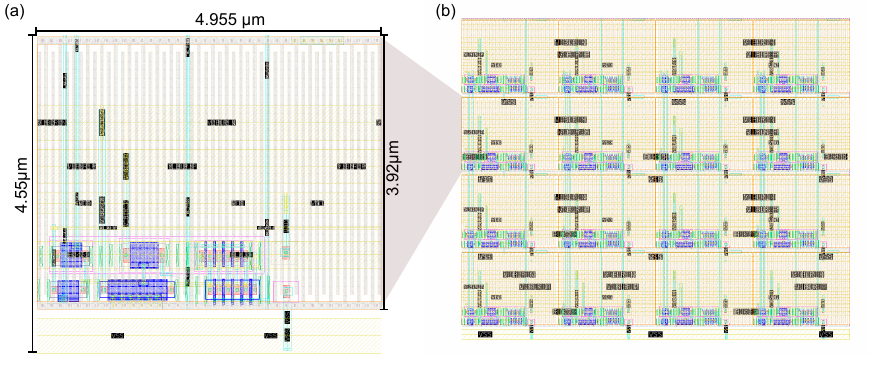}
    \caption{(a) Single cell layout of a for storing $V$ values and computing $\phi\left(S\right) \cdot V$. (b) $4 \times 4$ array Layout. Note that apart of the bottom row the hight scales with \valunit{3.92}{\mu m}}
    \label{fig:CMOS_layout}
\end{figure}

We design a custom CMOS layout of the proposed gain cell and charge-to-pulse circuits. In this study, the circuit simulations were done in TSMC \valunit{28}{nm} silicon CMOS technology. We used this conventional design style as a proof of concept to demonstrate the capacity of our gain cells-based architecture to perform the attention mechanism. However, CMOS gain cells lead to relatively large area footprint, primarily due to Metal-Oxide-Metal (MOM) capacitors which must be relatively large due to their high leakages. Our layout results show that each cell has a dimensions of \valunit{3.9}{\mu m}$\times$\valunit{4.9}{\mu m}, resulting in an area of \valunit{0.08}{mm^2} per $64\times64$ array, or \valunit{1.28}{mm^2} for one entire attention head (16 sub-tiles). In comparison, the ReLU charge-to-pulse circuitry and its signed variant occupy an area of \valunit{0.01}{mm^2} and \valunit{0.02}{mm^2} per attention head, respectively. The Layout of the gain cells storing $V$ values and computing $\phi\left(S\right) \cdot V$ is shown in shown in Fig.~\ref{fig:CMOS_layout}. Note that the Layout of the gain cells storing $K$ has transposed World Lines (WL) and Bit Lines (BL). 

\subsection{ReLU charge-to-pulse converter}
\label{supp:ReLU}
In this section, we provide additional information on the working principle for the ReLU charge-to-pulse circuit block. This charge-to-pulse circuit operates in three distinct phases: sampling, discharge, and reset. During the sampling phase, input pulses are applied to the first gain cell array, and the currents generated by the cells are integrated by a capacitor ($C_2$) in the charge-to-pulse circuit. This capacitor also utilizes the wire capacitance of the word line. In the discharge phase, the voltage of the capacitor $C_2$ is discharged with a constant current controlled by the bias voltage $V_b$. However it is important to note that the system employs an energy saving scheme by checking the voltage on the integrating capacitor $V_{cap}$ and only preforming the discharge in case the voltage is positive. An inverter acts as a simple comparator, triggering a pulse of variable width. Finally, in the reset phase, the bit line is reset to the initial bit line voltage to prepare for a new inference step.

\begin{figure}[htb!]
    \centering
    \includegraphics[width=1\linewidth]{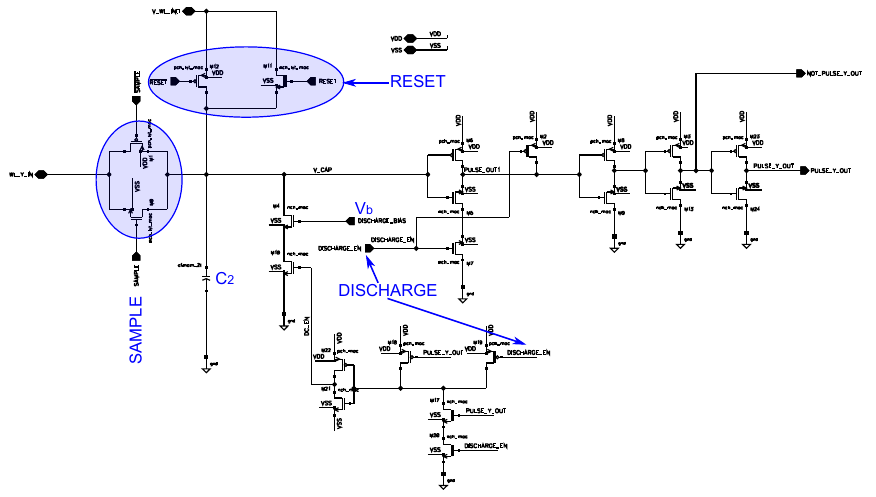}
    \caption{CMOS schematic of the ReLU charge-to-pulse converter}
    \label{fig:ReLU_c2p}
\end{figure}

\section{Signed charge-to-pulse converter}
\label{supp:signed}
\begin{figure}[htb!]
    \centering
    \includegraphics[width=1\linewidth]{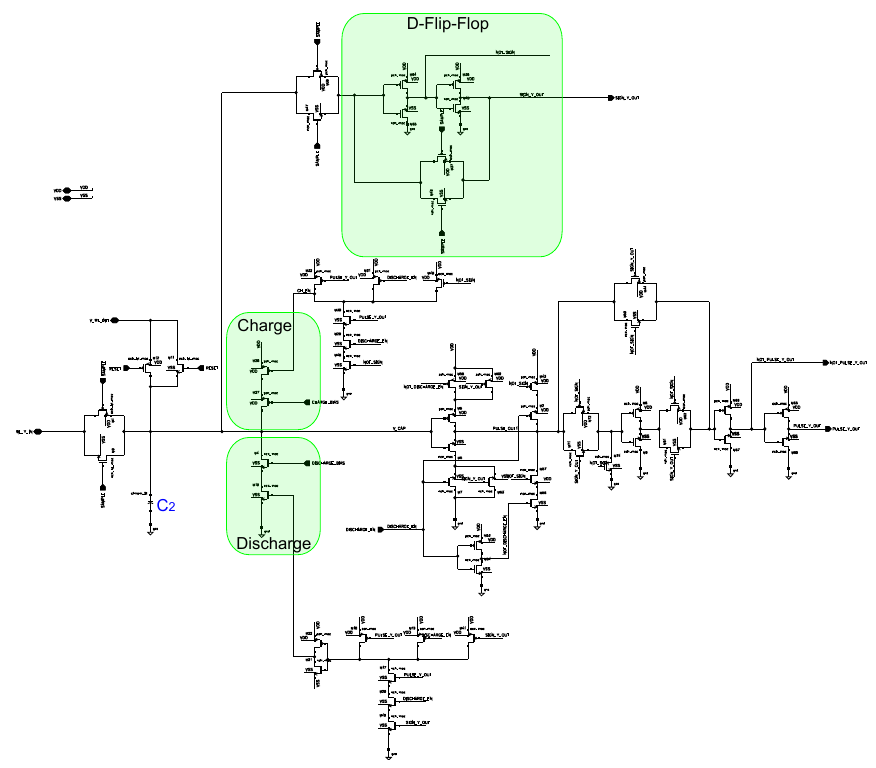}
    \caption{CMOS schematic of the signed charge-to-pulse converter}
    \label{fig:signedc2p}
\end{figure}
\begin{figure}[htb!]
    \centering
    \includegraphics[width=1\linewidth]{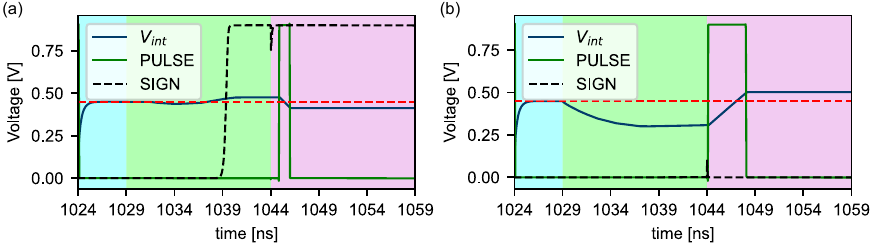}
    \caption{Example for a MAC result with positive signed (a) and negative sign (b) featuring distinct charging behaviours.}
    \label{fig:signed_c2p_example}
\end{figure}

To implement a signed charge to pulse circuit, the main difference from the circuit in \ref{supp:ReLU} is the addition of a charge-up path and a D Flip-Flop. This  serves the following purpose: at the end of the sampling stage, the D Flip-Flop captures the polarity of the voltage on the capacitor and stores it for subsequent operations. This stored sign determines whether a charge or discharge is applied to the capacitor voltage. The pulse-forming circuit is now slightly more complex to ensure consistent, high-active output pulses. Ultimately, the circuit outputs both the sign and the output pulses. Fig.~\ref{fig:signedc2p} shows the circuit schematics. The two distinct charge up and charge down behaviours given a certain SIGN are displayed in Fig.~\ref{fig:signed_c2p_example}.

\section{Adaptation Algorithm Generalization to Different Nonlinear Functions}
\label{supp:non_linear_adapt}
\begin{figure}[htb!]
    \centering    \includegraphics[width=\linewidth]{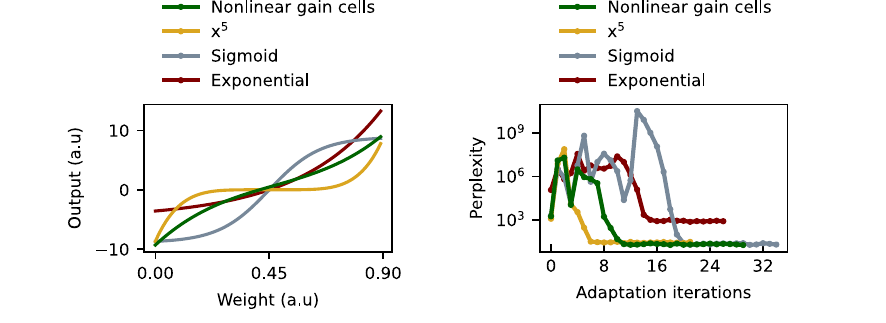}
    \caption{(a) Different nonlinear functions tested in place of the gain cells nonlinearities. (b) Evolution of perplexity (lower the better) during the adaptation algorithm when our attention model is implemented with different nonlinearities applied on the stored keys and values.}
    \label{fig:multi_nonlinear}
\end{figure}

In this experiment, we evaluate the capacity of our adaptation algorithm to generalize to other nonlinearities than the one modelling the gain cells (see Fig.~\ref{fig:multi_nonlinear}). We perform the dot-products of the attention mechanism with different nonlinearities applied to the stored keys and values. The different functions tested are: $f\left( x \right) = \alpha \left( x-\beta\right)^5$, $f\left( x \right) = \alpha sigmoid\left(10\left( x-\beta \right)\right)$, and $f\left( x \right) = \alpha e^{3\left( x-\beta \right)}$, with $\alpha$ and $\beta$ chosen to yield to similar ranges for the different functions. 

We see that our adaptation algorithm manage to reduce the perplexity drastically, except for the exponential function. The high asymmetry of the exponential function prevents the network to yield good accuracies. The adaptation algorithm manage to reduce the perplexity for functions which are anti-symmetrical even if they are highly nonlinear, such as $x^5$ (perplexity=29) or $sigmoid$ (perplexity=21).